# Video-Based Hand Pose Estimation for Remote Assessment of Bradykinesia in Parkinson's Disease

Gabriela T. Acevedo Trebbau[1], Andrea Bandini[2], and Diego L. Guarin[1]

[1]Department of Applied Physiology and Kinesiology, University of Florida, Gainesville, FL 32611 U.S.A
[2]Interdisciplinary Research Center "Health Science" – Scuola Superiore Sant'Anna, Pisa, Italy
d.guarinlopez@ufl.edu

**Abstract.** There is a growing interest in using pose estimation algorithms for video-based assessment of Bradykinesia in Parkinson's Disease (PD) to facilitate remote disease assessment and monitoring. However, the accuracy of pose estimation algorithms in videos recorded from video streaming services during Telehealth appointments has not been studied. In this study, we used seven off-the-shelf hand pose estimation models to estimate the movement of the thumb and index fingers in videos of the finger-tapping (FT) test recorded from Healthy Controls (HC) and participants with PD and under two different conditions: *streaming* (videos recorded during a live Zoom meeting) and *on-device* (videos recorded locally with high-quality cameras). The accuracy and reliability of the models were estimated by comparing the models' output with manual results. Three of the seven models demonstrated good accuracy for *on-device* recordings, and the accuracy decreased significantly for *streaming* recordings. We observed a negative correlation between movement speed and the model's accuracy for the *streaming* recordings. Additionally, we evaluated the reliability of ten movement features related to bradykinesia extracted from video recordings of PD patients performing the FT test. While most of the features demonstrated excellent reliability for *on-device* recordings, most of the features demonstrated poor to moderate reliability for *streaming* recordings. Our findings highlight the limitations of pose estimation algorithms when applied to video recordings obtained during Telehealth visits, and demonstrate that *on-device* recordings can be used for automatic video-assessment of bradykinesia in PD.

## 1    Introduction

There is a growing interest in developing methods for automatic, video-based quantification of motor symptoms in Parkinson's Disease; in particular, many studies have proposed methodologies for assessing bradykinesia from videos of the finger tapping (FT) test. [1]–[4]. Bradykinesia is a cardinal motor symptom of PD characterized by slowness of movement, decrease in movement amplitude, hesitations/halts during movement, and progressive decrease in movement speed [5], [6]. The FT test is a clinical motor assessment used to evaluate upper-limb bradykinesia, and many studies have developed machine learning algorithms to assess bradykinesia severity using videos of the FT tests. Most of these studies use markerless hand pose estimation algorithms to estimate the subjects' hand movements during the FT test and quantify the degree of bradykinesia [1]–[4], [7]–[10].

Available methods automatically assess bradykinesia from standard videos of the FT test, making them potentially useful tools to support remote assessment during Telehealth appointments. However, most studies employ high-quality videos recorded in controlled laboratory or clinical settings where technical components related to video quality, frame rate, and image resolution were tightly controlled throughout the recordings. Such conditions are not common during Telehealth appointments as Telehealth platforms prioritize connection latency over image quality and frame rate [11]. It is not known to what extent aspects related to image quality, information loss, and variable frame rate affect the accuracy of markerless hand pose estimation algorithms and negatively impact the automatic assessment of bradykinesia. There is a need to validate the accuracy of hand pose estimation algorithms in videos recorded under conditions similar to those encountered in Telehealth appointments.

This study aims to evaluate the accuracy and reliability of automatically estimated hand movements from videos of the FT test recorded during two conditions: (1) *streaming* (i.e., Zoom recordings), and (2) *on-device* (high-quality recordings). We performed two experiments to explore the effects of the recording conditions on the accuracy and reliability of hand pose estimation algorithms. For the first experiment, we investigated the accuracy of seven off-the-shelf hand pose estimation algorithms for tracking hand movements during the FT test in healthy subjects for both recording conditions. We compared the results provided by the algorithms against manual annotations. For this experiment, we employed videos of healthy controls demonstrating a wide range of movement speeds and amplitudes. For the second experiment, we investigated the reliability of video-based kinematic features related to bradykinesia derived from videos of subjects previously diagnosed with PD and under stable deep brain stimulation (DBS).

### 1.1 Previous Work

Multiple studies have used machine learning algorithms for automatic assessment of bradykinesia from videos of the FT test. A common approach involves using hand pose estimation algorithms to track the movement of the index and thumb fingers during the test, and then compute movement and velocity-based features that can be used to identify persons with PD from healthy controls and to estimate disease severity by predicting a clinical score such as the Movement Disorder Society – Unified Parkinson's Disease Rating Scale (MDS-UPDRS) score for the FT test (ranging from 0 to 4). H. Li et al. proposed a method that estimates hand pose features (location of 21 joints in the hand), motion features (temporal inter-frame variation of the hand skeleton data), and geometry features (inter-joint relationship of hand skeleton data) to automatically classify the motor severity based on the MDS-UPDRS score for the FT test [1]. Their proposed method for automatic scoring achieved an accuracy of 72.4%, demonstrating accurate and reliable results [1]. Z. Li, et al. proposed using one-dimensional time series data of velocity and movement range of the finger movements as input features to predict the MDS-UPDRS score for FT [2]. Using a fivefold cross-validation method, they achieved an average prediction accuracy of 79.7% [2]. In addition, they analyzed each feature set individually to investigate the importance of range and velocity information for predicting the MDS-UPDRS score and found that the accuracy obtained using the movement range data was higher than the accuracy obtained using the velocity data (68.4% and 46.8%, respectively) [2]. N. Yang, et al. proposed using tapping rate, tapping frozen times, and tapping amplitude variation as input features to predict the FT test severity [4]. They evaluated their method for each hand separately; for the left hand they obtained a micro average of precision, recall, and f1-score of 88%, 89%, and 88%, respectively, and for the right hand they obtained a micro average of precision, recall, and f1-score of 85%, 85%, and 84%, respectively [4]. Y. Liu, et al., proposed using four parameters related to finger movement amplitude and velocity, and their variabilities to automatically predict the MDS-UPDRS score for the FT test [7]. They achieved an average accuracy of 89.7% [7]. K. W. Park, et al., employed pose estimation algorithms to extract features related to speed, amplitude, and fatigue during the FT task. They obtained good agreement (Intra Class Correlation = 0.8) and an absolute agreement rate of 70% between the predicted and the clinician scores [10]. Finally, G. Morinan et al. proposed a computer-vision-based approach to extract features relevant to bradykinesia and predict the MDS-UPDRS ratings for the limb-based bradykinesia items [9]. The features proposed described the main aspects evaluated during the MDS-UPDRS; amplitude, speed, hesitations and halts, and decrementing amplitude and speed [9]. They obtained an acceptable accuracy (the percentage of estimates for which the prediction error was zero or +/-1) of 84%, for the prediction of the MDS-UPDRS score for the FT test [9].

## 2 Methods

### 2.1 Participants

**Experiment 1.** Ten healthy subjects participated in this study (6 females, 4 males; age range 19-57). Participants were included if they: (1) were fluent in English; (2) have no self-reported history of neurological or movement disorders; and (3) did not demonstrate any signs of cognitive impairments as measured by a score of 26 or higher in the Montreal Cognitive Assessment (MoCA) [12]. All participants were recorded during one session at Florida Institute of Technology in Melbourne, Florida, U.S. The study was approved by the institutional Research Ethics Board and participants signed an informed consent form according to the declaration of Helsinki.

**Experiment 2.** Six patients previously diagnosed with Parkinson's Disease (PD) participated in this study (1 female, 5 males; age range 64-71). Participants were included if they: (1) were fluent in English; (2) have been previously diagnosed with Parkinson's Disease by a movement disorders specialist using the UK PD Brain Bank diagnostic criteria [13]; (3) were under stable DBS; and (4) did not demonstrate signs of moderate cognitive impairments as measured by a score of 18 or higher in the MoCA [14]. All participants were recorded during one session at the University of Florida, Gainesville, Florida, U.S. Participants were in the OFF-medication state, they had not taken any medication to control their Parkinsonism for at least 12h before starting the recording. The study was approved by the institutional Research Ethics Board and Participants signed an informed consent form according to the declaration of Helsinki.

## 2.2 Recordings

**Experiment 1.** The recording sessions were held in a quiet, well-illuminated room. Participants sat down comfortably facing a Logitech BRIO camera positioned on a tripod and the webcam of a Dell XPS laptop positioned on a table. A Zoom meeting was held between the laptop and a computer workstation set up in an office nearby the recording room. The workstation was connected to the internet through ethernet, and the laptop was connected to the local Wi-Fi network. The camera and laptop's webcam were adjusted to record the same view, which included the subjects' torso and head. One researcher was in the recording room guiding the participants, while another researcher was supervising the Zoom meeting. We acquired videos using two recording setups: (1) high-speed Logitech BRIO camera, recording at 100 fps with a resolution of 1280x720 pixels (px) (*on-device*); and (2) built-in webcam from a Dell XPS laptop computer streaming through a Zoom meeting (*streaming*), the Zoom meeting was recorded at 25 fps with a resolution of 640x360 px on the computer workstation.

During the recording session, participants were asked to tap their index and thumb as fast as possible, fully separating the fingers after each tap for fifteen seconds. Subjects were requested to start the task with the fingers fully separated. We collected a total of four videos from each participant: one video for each hand (right and left) and with each camera (on-*device* and *streaming*).

**Experiment 2.** The recording sessions were held in a well-illuminated clinical examination room. Participants sat down facing an iPhone 12 and an iPad that were set up on two tripods. A Zoom meeting was held between the iPad and a computer workstation set up in an office nearby to the recording room. The workstation, iPad, and iPhone were connected to the local Wi-Fi network. The iPhone and iPad cameras were adjusted to record the same view, which included the participant's whole body. One researcher was in the examination room guiding the participants, while another researcher was supervising the Zoom meeting. We acquired videos using two recording setups: (1) iPhone 12 camera, recording at 60 fps with a resolution of 1080x1920 px (*on-device*); and (2) iPad front camera, streaming through a Zoom meeting (*streaming*), the Zoom meeting was recorded at 25 fps with a resolution of 640x360 px on the computer workstation.

Participants with PD arrived at the recording room with the Deep Brain Stimulators (DBS) turned ON and performed an initial recording session of the FT test (DBS ON). Then, the DBS was turned OFF and the recording session was repeated (DBS OFF). A total of eight videos were collected from each participant with PD, one video for each hand (right and left), each camera (*on-device* and *streaming*) and with the DBS turned ON and OFF (DBS ON and DBS OFF).

### 2.3    Video Analysis

**Manual Annotations.** A single annotator manually localized the (x, y) coordinates of the tip of the thumb and index fingers and calculated the Euclidian distance in each video frame, resulting in a distance time series with the same length as the video. The distance signals were smoothed using a Savitzky-Golay filter and normalized between 0 and 1, indicating the minimum and maximum distance between the fingers. These signals were considered as the ground truth for the remainder of the study.

**Pose Estimation.**

*Experiment 1.* **Table 1** describes the seven off-the-shelf hand pose estimation models used in this study. The open-source models are available in two Python libraries: MediaPipe and MMPose. The models' input was a video frame containing one or two hands, and the output was the (x, y) coordinates of 21 hand landmarks for each hand, corresponding to the base of the hand, the carpometacarpal, metacarpophalangeal, and interphalangeal joints, and tip of the thumb, and the metacarpophalangeal, proximal interphalangeal, and distal interphalangeal joins, and tip of the remainder fingers. We extracted the (x, y) coordinates for the tip of the thumb and index fingers and calculated the Euclidean distance between these two landmarks for each video frame, resulting in a distance time series with the same duration as the video recording. Signals were smoothed and normalized as the ground truth signals.

*Experiment 2.* We selected the best-performing model from the results of Experiment 1 to analyze the video from participants with PD. We followed the same protocol as Experiment 1 to extract, smooth, and normalize the distance signals.

**Feature Extraction.** From the distance signals obtained in Experiment 2, we extracted ten movement features related to bradykinesia, mean and coefficient of variation (cv) of movement frequency, mean and cv of movement amplitude, mean and cv of movement speed, range of period duration, roughness, decrement in amplitude, and decrement in speed. These features have been used previously for video-based assessment of Bradykinesia [9]. These features are based on the peaks (maximum opening) and valleys (maximum closing) of the distance signal; we identify the peaks and valleys manually from the videos to guarantee consistency among the different recording environments. A description of the features is presented in the supplementary material. Features extracted from the manually annotated, *on-device* videos were considered as the ground truth for the remainder of this study.

**Table 1.** Description of the Seven Hand Pose Estimation Models Used in this Study.

| Model Name | Python Library | Training Database | Architecture |
|---|---|---|---|
| *model 1* | MediaPipe | Google's Database | BlazePose |
| *model 2* | | | HRNet [20] |
| *model 3* | | COCO Hands [19] | MobileNet [22] |
| *model 4* | MMPose | | ResNet [21] |
| *model 5* | | | HRNet [20] |
| *model 6* | | OneHand10k [18] | MobileNet [22] |
| *model 7* | | | ResNet [21] |

### 2.4 Statistical Analyses

**Experiment 1.** We calculated the coefficient of determination ($R^2$ score) to evaluate the similarity between the ground truth and automatically estimated distance signals. We performed the Shapiro-Wilk test of the $R^2$ scores to test for normality and compared $R^2$ scores obtained from the *on-device* and *streaming* recordings using a t-test or a Mann-Whitney U test in case of non-normal distributions. We also performed a Spearman Correlation analysis between the subjects' maximum speed and $R^2$ score. The level of statistical significance was set at 0.05.

**Experiment 2.** We calculated the Intraclass correlation coefficient (ICC (2,1); single random rater, absolute agreement) to assess the reliability between the ground truth features and the features estimated using a hand pose estimation algorithm. Values with negative ICC were set to 0.0 as suggested by J. J. Bartko [15]. ICC values <0.50 indicated poor reliability, values between 0.50 to 0.74 indicated moderate reliability, values between 0.75 to 0.9 indicated good reliability, and values >0.9 indicated excellent reliability [16]. The level of statistical significance was set at 0.05.

## 3 Results

### 3.1 Experiment 1: Pose estimation for automatic movement tracking.

**Figure 1** presents two frames from the *streaming* and *on-device* recordings from a HC demonstrating the maximum opening and closing of the fingers during the FT test. **Figure 2** presents the manually and automatically derived distance signals for the right hand of two HC demonstrating the lowest and highest movement speeds.

**Table 2** shows the results of the statistical analysis comparing the $R^2$ score between the manual and automatically derived distance signals for the *on-device* and *streaming* recordings. *Model 1, model 5,* and *model 7* achieved the highest average $R^2$ score for *streaming* and *on-device* recordings. All other models demonstrated $R^2$ score values lower than 0.9 for *streaming* and *on-device* recordings, with some models demonstrating negative $R^2$ score values. The last column of **Table 2** presents the statistical comparison between the $R^2$ score obtained for *on-device* and *streaming* recordings. For most models, the *on-device* recordings demonstrated significantly higher $R^2$ scores when compared to the *streaming* recordings. *Model 3* and *model 4* did not show significantly different $R^2$ scores between the recording conditions.

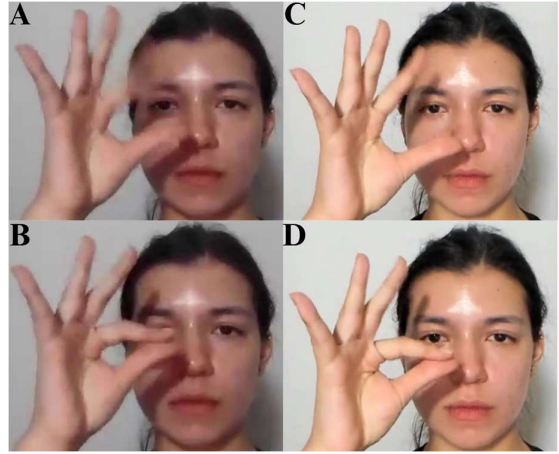

**Fig. 1.** Frames from a video recording of the FT test. A and B are frames from the *streaming* recording A) maximum opening (peak), and B) maximum closing (valley). C and D are frames from the *on-device* recording, C) maximum opening (peak), and D) maximum closing (valley)

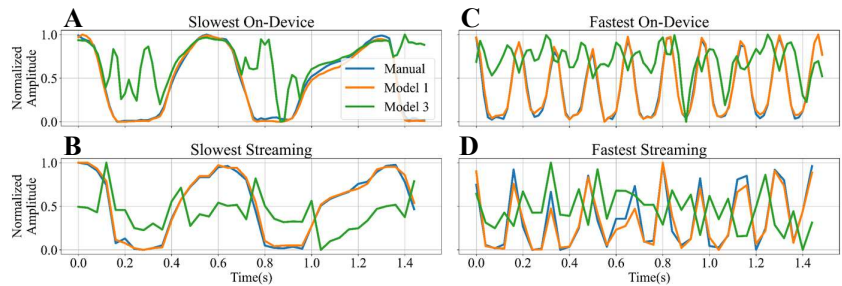

**Fig. 2.** Blue lines show the manually derived distance signals between the tip of the index and thumb fingers of the right hand for the HC participants with the lowest (left column, maximum speed of 8.35/s) and highest (right column, maximum speed of 22.3/s) maximum speeds. Orange and green lines show the distance signals between the tip of the index and thumb fingers yielded by the best and worst performing models (Model 1 and Model 3, respectively). First and second row present the results obtained with *on-device* and *streaming* recordings respectively.

**Fig. 3** shows the results obtained for the correlation analysis between $R^2$ scores and the maximum speed for the best performing models. For the *streaming* recordings, there was a significant negative correlation between the subjects' maximum speeds and $R^2$ scores for *models 1, 5,* and *7*. The correlation results and respective p-values are presented in the supplementary materials.

**Table 2.** Statistical Analyses Results for Experiment 1

| Model | $R^2$ score Mean ± Std | | P-value |
|---|---|---|---|
| | **Streaming** | **On-Device** | |
| *model 1* | 0.90 ± 0.10 | 0.98 ± 0.02 | <0.001[A] |
| *model 2* | 0.23 ± 0.78 | 0.73 ± 0.44 | 0.005[A] |
| *model 3* | -0.78 ± 0.91 | -0.32 ± 0.84 | 0.105[B] |
| *model 4* | -0.17 ± 1.14 | 0.31 ± 0.66 | 0.102[A] |
| *model 5* | 0.73 ± 0.25 | 0.97 ± 0.04 | <0.001[A] |
| *model 6* | 0.53 ± 0.52 | 0.88 ± 0.19 | 0.002[A] |
| *model 7* | 0.67 ± 0.36 | 0.95 ± 0.11 | <0.001[A] |

[A] Mann Whitney U Test
[B] T-test

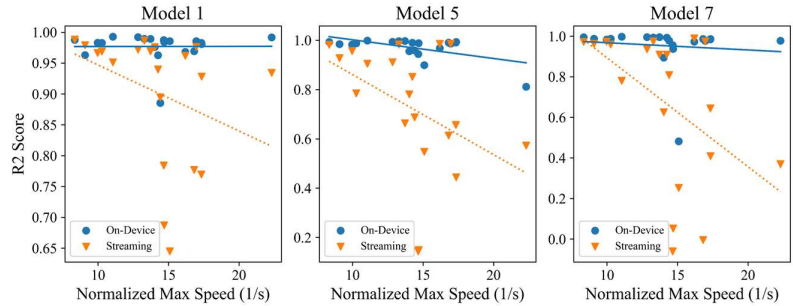

**Fig. 3.** Scatter plot showing the relation between maximum speed and $R^2$ scores for *model 1, 5* and *7*. The blue and orange results represent the *on-device* and *streaming* recordings, respectively. The lines represent the fest fit results obtained via least mean squares minimization.

### 3.2 Experiment 2: Reliability of automatic pose estimation-based features.

***On-device* Recordings:** For the *on-device* recordings (see supplemental material), most of the features achieved excellent reliability for DBS ON and DBS OFF (ICC > 0.90). Only *Roughness* demonstrated poor reliability for DBS OFF (ICC = 0.37) and moderate reliability for DBS ON (ICC = 0.72).

***Streaming* Recordings:** For the *streaming* recordings (see supplemental material), most features achieved poor or moderate reliability for DBS ON and DBS OFF (ICC from 0.0 to 0.71). Only measures related to movement frequency and period duration achieved good or excellent reliability (ICC => 75).

## 4    Discussion

In this study, we used hand pose estimation algorithms to automatically track the movements of the hand and extract movement features from videos of the FT test, a widely used motor task for assessment of upper limbs bradykinesia in PD. We explored the effects of various recording conditions, including *on-device* and *streaming* recordings, on the accuracy of pose estimation models. The *streaming* recordings are similar to those observed during Telehealth appointments, so that our results are relevant to understand the limitations of hand pose estimation algorithm in the remote assessment of bradykinesia. Additionally, we explored the impact of both recording conditions on the reliability of ten video-based movement features related to bradykinesia.

### 4.1 Accuracy of hand pose estimation models

The first part of this study evaluated the performance of different hand pose estimation models during the FT test. Our results indicate that *model 1* is the best-performing model for pose estimation, demonstrating $R^2$ scores of at least 0.90 for all recording settings. This model is part of Google's MediaPipe and was trained with a private dataset consisting of 30.000 natural and artificial images [17]. Regarding the other models, we observed that models trained with the OneHand10k dataset demonstrated better performance compared to those trained with the COCO database. This observation might be explained because the OneHand10K dataset includes single hand images covering a wide range of hand poses [18], whereas the COCO dataset contains full body images [19]. Thus, models trained with the OneHand10K dataset are optimized for hand pose estimation, while the models trained with the COCO dataset are optimized for whole-body pose estimation. Moreover, HRNet outperformed all other architectures, achieving the highest R2 scores with the given dataset. This result might be attributed to HRNet's ability to maintain the frame's resolution throughout the process, resulting in more spatially precise representations [20]. In contrast, ResNet and MobileNet initially encode the input frame as low-resolution representations [21], [22], leading to reduced accuracy. Our results also demonstrated that *on-device* recordings yielded significantly higher $R^2$ scores than *streaming* recordings for most models. Zoom adjusts the true frame rate and image compression based on the internet connectivity, resulting in video with variable video quality and repeated frames. During the streaming recordings we observed images with high pixelization and blurring, especially during high-speed movement - see for example Figure 1A -, which negatively affected the accuracy of hand pose estimation algorithms. In contrast, *on-device* recordings have constant frame rate and compression throughout the video, resulting in overall high accuracy.

Our result also showed that the movement maximum speed significantly impacted the model's performance for the *streaming* recordings. We link this phenomenon with the blurring effect observed in the *streaming* recording, where the shape of the fingers was lost for high-speed movements, causing the hand pose estimation model to misplace the finger. These results are important because inconsistent rhythm and the sequence effect can appear early in PD, while slowness of movement will appear later as the disease progresses. Based on this observation, we argue that *streaming* recordings with variable video quality and rate are not adequate for assessing bradykinesia from videos of the FT test as assessment will be influenced by the movement speed.

### 4.2 Reliability of automatically estimated movement features

The second part of this study investigated the reliability of video-based movement features used to quantify bradykinesia across different recording conditions. When considering *on-device* recordings, most of the features resulted in good to excellent reliability. In contrast, *Roughness'* reliability was poor for DBS OFF (ICC = 0.37) and moderate for DBS ON (ICC = 0.72). The low reliability is likely because *Roughness* relies on acceleration and jerk, the second and third-order derivatives of the distance signal, and derivatives are known to amplify the signal's noise [23]. Our findings suggest that features based on the distance signal and its first-order derivative (velocity) can be reliably estimated from *on-device* videos recorded at 60 fps with a resolution of 1080x1920 px but features based on higher-order derivatives require different recording conditions to the ones explored in this study.

When considering *streaming* recordings, most of the features demonstrated poor to moderate reliability. Features that did not depend on the results provided by the hand pose estimation algorithm provided excellent or good reliability, including *MeanFreq*, *CovarFreq*, and *PeriodRange*. The low reliability can be explained by errors in the distance signal caused by the low video quality. These findings suggest that there is a clear impact of the streaming recordings' video quality in the estimation of pose estimation-based features and highlights the limitations of using streaming recordings for automatic assessment of bradykinesia from videos of the FT test.

### 4.3 Limitations

Several limitations should be considered when interpreting the results. First, the small sample size restricts the generalizability of the findings and limits the statistical power of the analyses. Future studies with larger sample sizes are needed to validate and strengthen our findings. Second, the setup for the streaming video recordings was standardized for all subjects, which does not capture the variations in connectivity, lighting conditions, and background noise commonly encountered in real-world Telehealth appointments. These variations could potentially impact the streaming quality and may influence the performance of pose estimation models. Future work should explore different streaming setups, proving a wider range of situations encountered in Telehealth visits. Third, the investigation focused solely on one streaming platform, Zoom, which does not represent the technical capacities and quality offered by other platforms. Exploring multiple streaming platforms would provide a broader perspective on the influence of streaming conditions in automatic movement tracking applications. Another limitation is that for this study we manually selected the peaks and valleys in the distance signal, which might not be efficient in real-world applications. Consequently, the impact of low-quality conditions was not appreciated for features that depend on movement frequency and period duration. Future works should include algorithms for the automatic detection of peaks and valleys to investigate the impacts of streaming conditions in fully automated systems for feature extraction.

## 5    Conclusion

In conclusion, this study focused on evaluating the accuracy and reliability of pose estimation-based movement data estimated from videos of the FT test recorded on-device and streaming settings. Our results demonstrated that Google MediaPipe and other off-the-shelf hand pose estimation algorithms provide accurate and reliable hand movement tracking and movement features for *on-device* recordings. These findings suggest that is feasible to use *on-device* recordings in combination with hand pose estimation algorithms for automatic, remote assessment of bradykinesia in PD. In contrast, the models' performance of the algorithms significantly decreased when applied to *streaming* video recordings. In particular, the model's performance was severely affected by the movement speed, with higher speed resulting in significantly worst performance. Moreover, clinically relevant movement features automatically estimated from streaming videos demonstrated poor reliability when compared to manually derived measures. These findings suggest that *streaming* recordings might not be adequate for automatic assessment of bradykinesia in PD.